\def\year{2020}\relax
\documentclass[letterpaper]{article}
\usepackage{aaai20}
\usepackage{times}
\usepackage{helvet}
\usepackage{courier}
\usepackage[hyphens]{url}
\usepackage{graphicx}
\usepackage{amsmath}
\title{Zero-Shot Learning from Adversarial Feature Residual to \\ Compact Visual Feature}
\author{
Bo Liu,\textsuperscript{\rm 1}\textsuperscript{\rm 2}
Qiulei Dong,\textsuperscript{\rm 1}\textsuperscript{\rm 2}\textsuperscript{\rm 3}\thanks{Corresponding author: Qiulei Dong}
Zhanyi Hu\textsuperscript{\rm 1}\textsuperscript{\rm 2}\textsuperscript{\rm 3}\\
\textsuperscript{\rm 1}National Laboratory of Pattern Recognition, Institute of Automation, Chinese Academy of Sciences\\
\textsuperscript{\rm 2}University of Chinese Academy of Sciences\\
\textsuperscript{\rm 3}Center for Excellence in Brain Science and Intelligence Technology, Chinese Academy of Sciences\\
bo.liu@nlpr.ia.ac.cn, qldong@nlpr.ia.ac.cn, huzy@nlpr.ia.ac.cn
}
\begin{document}
\maketitle
\begin{abstract}
Recently, many zero-shot learning (ZSL) methods focused on learning discriminative object features in an embedding feature space, however, the distributions of the unseen-class features learned by these methods are prone to be partly overlapped, resulting in inaccurate object recognition. Addressing this problem, we propose a novel adversarial network to synthesize compact semantic visual features for ZSL, consisting of a residual generator, a prototype predictor, and a discriminator. The residual generator is to generate the visual feature residual, which is integrated with a visual prototype predicted via the prototype predictor for synthesizing the visual feature. The discriminator is to distinguish the synthetic visual features from the real ones extracted from an existing categorization CNN. Since the generated residuals are generally numerically much smaller than the distances among all the prototypes, the distributions of the unseen-class features synthesized by the proposed network are less overlapped. In addition, considering that the visual features from categorization CNNs are generally inconsistent with their semantic features, a simple feature selection strategy is introduced for extracting more compact semantic visual features. Extensive experimental results on six benchmark datasets demonstrate that our method could achieve a significantly better performance than existing state-of-the-art methods by $\sim$$1.2$-$13.2\%$ in most cases.
\end{abstract}

\begin{figure}[t]
\centering
\includegraphics[width=0.8\columnwidth]{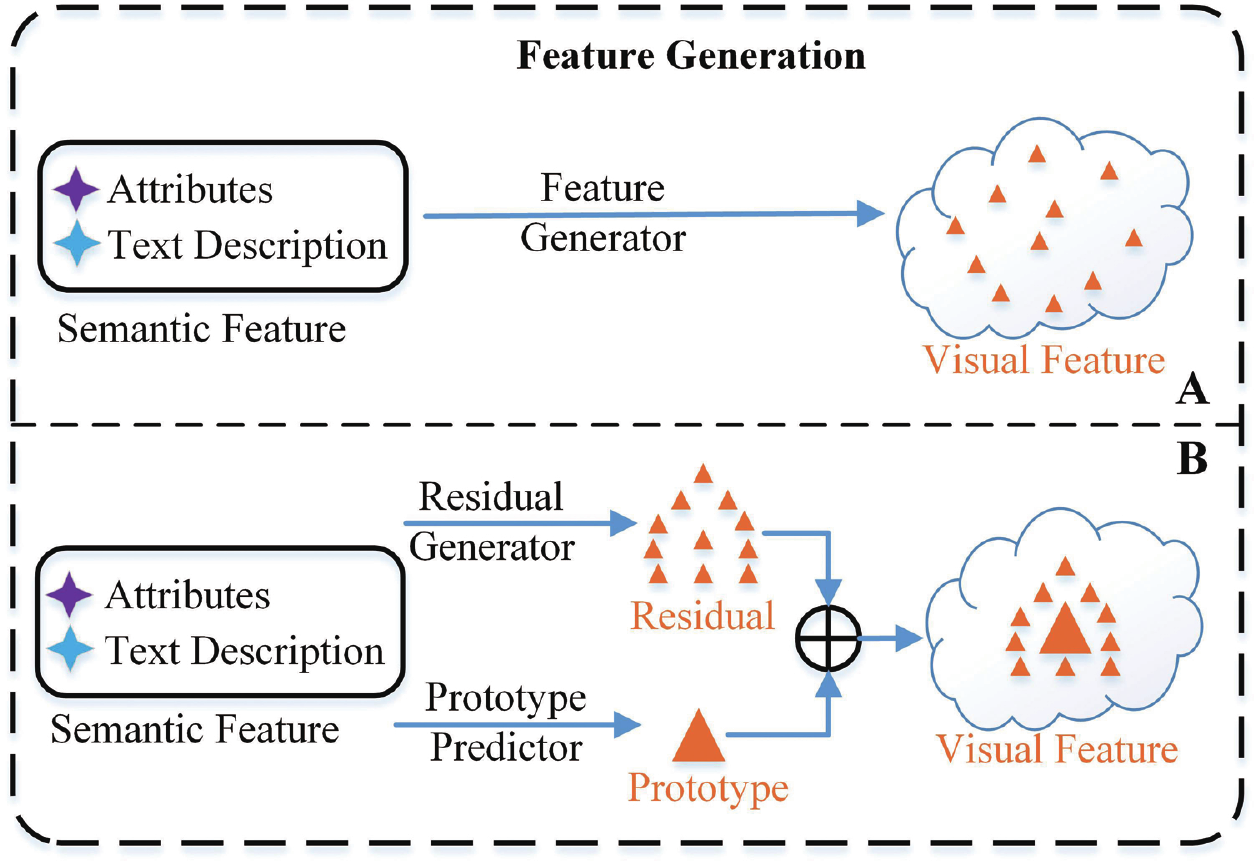}
\caption{Comparison of our method with existing GAN-based ZSL methods. A: Existing GAN-based methods generate visual features conditioned on their semantic feature. B: Our method generates visual feature residuals conditioned on their semantic feature, and then synthesizes visual features by combining the residuals and a visual prototype predicted from its semantic feature.}
\end{figure}

\section{Introduction}
In recent years, zero-shot learning (ZSL) has attracted more and more attention in pattern recognition and machine learning. Given a set of labeled seen-class data as well as the semantic relationship between seen and unseen classes, ZSL aims to recognize unseen-class instances. Most existing ZSL methods focused on learning discriminative object features in an embedding feature space where the semantic relationship between seen and unseen classes is preserved, and they could be roughly divided into two categories: visual-to-semantic methods and semantic-to-visual methods.

The visual-to-semantic methods \cite{Frome13DeViSE,Akata16ALE,xian2016LATEM} aim to build a projection function from visual features to semantic features. The visual features are generally extracted from the input images by CNNs (Convolutional Neural Networks), while the semantic features describe semantic attributes of object class, e.g. class-level attribute and text description. The projection function is trained on seen-class data by making the projected visual features closer to the semantic feature of their correct class. However, the distributions of the projected visual features of unseen classes by these methods are prone to be partly overlapped, leading to inaccurate object recognition.

In comparison to these visual-to-semantic methods, the semantic-to-visual methods \cite{Zhu18GAZSL,Xian18FCLSWGAN,li2019LiGAN} have significantly improved the ZSL performance recently. Most of semantic-to-visual methods aim to generate visual features of unseen classes conditioned on their semantic features via generative adversarial network (GAN) as illustrated in Figure 1 A, and then train a classifier with the synthetic visual features and their corresponding labels for classifying real visual features of unseen classes. Despite their success, these GAN-based methods are still limited by the problem that the distributions of the synthetic unseen-class visual features are partly overlapped. Figure 2 A provides an example for illustrating this problem. As shown in Figure 2 A, different color points represent the visual features belonging to different unseen classes, which are generated by an existing GAN-based ZSL method (e.g. \cite{Xian18FCLSWGAN}). Obviously, there is some overlap between different color points, indicating that the distributions of the synthetic visual features are partly overlapped. This overlap is probably because the GAN used to generate the unseen-class visual features is trained only on the seen-class data.

In addition, the fidelity of synthetic visual features of unseen classes by these GAN-based methods is also limited by the inconsistency between semantic features and visual features. The semantic-visual inconsistency refers to the fact that even if two classes have very similar semantic attributes, e.g. both elephants and tigers have the `tail' attribute, their visual features could be very different, e.g. the visual features of an elephant's tail and a tiger's tail are quite different. As illustrated in Figure 2 C, due to this inconsistency, even though GAN can learn an accurate semantic-to-visual generative relationship (illustrated by straight line for convenience) on seen-class data, the distribution of synthetic unseen-class visual features (illustrated by bigger ellipse) according to the learned generative relationship is different from the distribution of real unseen-class visual features.

Addressing these two problems, we propose a novel adversarial network to learn compact semantic visual features for ZSL by integrating the visual prototype and the visual feature residual, which consists of a residual generator, a prototype predictor, and a discriminator. Here, the visual prototype represents general visual features of each class and the visual feature residual represents the feature deviation of each sample from its prototype. The residual generator is employed to generate the visual feature residual conditioned on semantic feature, and then the visual feature is synthesized by combining the residual with the class-level visual prototype predicted from its semantic feature by the prototype predictor, as illustrated in Figure 1 B. After the synthetic visual features are synthesized, the discriminator tries to distinguish the synthetic visual features from the real ones extracted from an existing categorization CNN. Since the visual prototypes are explicitly predicted for both seen and unseen classes and most of the residuals are generally numerically much smaller than the distances among prototypes of all classes, the synthetic visual features by the proposed method are less overlapped as shown in Figure 2 B. To alleviate the semantic-visual inconsistency problem, we propose a simple feature selection strategy which is able to adaptively select some semantically consistent feature dimensions from the original visual feature.

In summary, our contributions are three-fold:
\begin{itemize}
\item We propose a novel adversarial network to learn compact semantic visual features for ZSL, which are synthesized by integrating the generated feature residuals and predicted visual prototypes. The distributions of the synthetic visual features by the proposed method are less overlapped. To our best knowledge, this is the first work to utilize adversarial feature residual for ZSL.
\item We propose a simple feature selection strategy that is able to adaptively select semantically consistent visual feature elements from the original visual feature, alleviating the semantic-visual inconsistency problem to some extent.
\item Extensive experimental results demonstrate that the proposed method can outperform existing state-of-the-art methods with a significant improvement on six benchmark datasets.
\end{itemize}
\begin{figure}[t]
\centering
\includegraphics[width=0.8 \columnwidth]{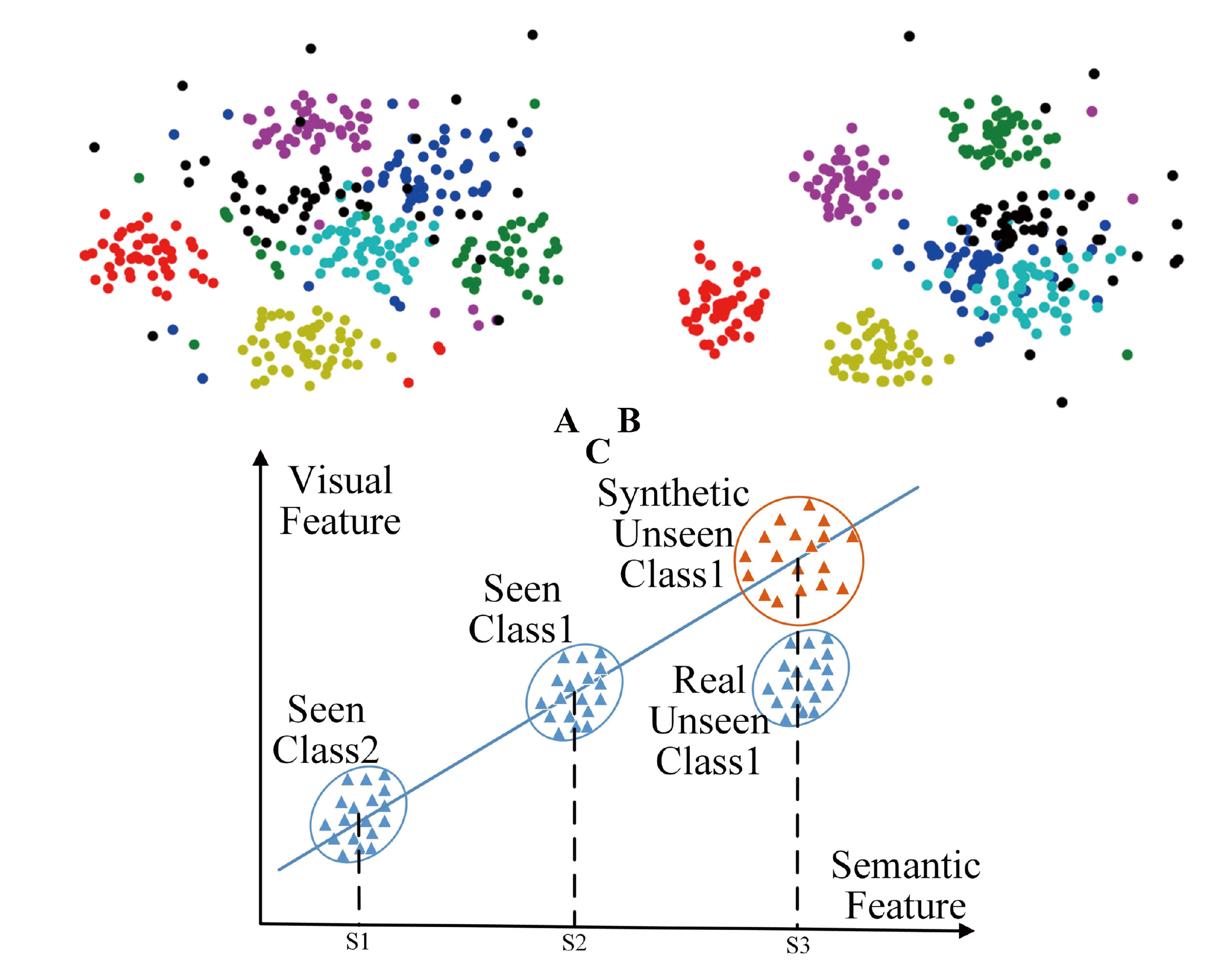}
\caption{A: t-SNE visualization of synthetic unseen-class visual features by an existing GAN-based ZSL method, different color points represent the visual features belonging to different unseen classes. B: t-SNE visualization of synthetic unseen-class visual features by our method. C: A ellipse represents a distribution of the visual features belonging to a class, the distribution of the synthetic visual features of the unseen class (bigger ellipse) according to the semantic-visual generative relationship (straight line) learned on two seen classes is different from the real counterpart.}
\end{figure}

\section{Related Work}
\textbf{Zero-Shot Learning.} Lampert et al. \cite{Lampert14DAP} proposed a two-step attribute-based classification method, where a probabilistic classifier was firstly learned for predicting probability of each attribute for each image, then the image was classified by a Bayesian classifier based on probabilities of attributes. Frome et al. \cite{Frome13DeViSE} proposed an end-to-end visual-to-semantic projection method. In this method, visual features extracted by an categorization CNN were projected into a semantic feature space by a linear function which was trained to make the projected visual features closer to the semantic feature of their correct category. Following this work, a lot of methods have devoted themselves to improve it by replacing its loss function \cite{akata2015sje,Akata16ALE,romera2015ESZSL,norouzi2013CONSE} or using nonlinear projection function \cite{socher2013CMT,xian2016LATEM}. Recently, some works proposed to learn a semantic-to-visual mapping, they \cite{Changpinyo17EXEM,zhang2017DEM} used semantic features to predict visual features by a transformation function, or they \cite{Zhu18GAZSL,Xian18FCLSWGAN} trained a GAN \cite{goodfellow2014gan} to generate visual features of unseen classes conditioned on their semantic features. Except for these, some works which leveraged mutually visual-semantic reconstruction \cite{Kodirov17SAE} or projected semantic features to parameter space \cite{Changpinyo16SYNC} were also proposed.\\
\textbf{Visual Prototype Prediction.} Visual features have a clustered structure in feature space, so it is feasible and beneficial to adopt a visual prototype to represent a class. Only a few works have applied visual prototype prediction to ZSL. Changpinyo et al. \cite{Changpinyo17EXEM} proposed a visual exemplar prediction method, where they trained a prediction function from semantic embeddings to visual exemplars and then the predicted exemplars were applied to other methods as visual training data or ideal semantic embeddings. \\
\textbf{Visual Feature Generation.} With the development of GAN, some works \cite{Zhu18GAZSL,Xian18FCLSWGAN,li2019LiGAN,paul2019SABR} have applied GAN to ZSL problem. In these methods, they all employed GAN to generate visual features of unseen classes conditioned on semantic features and then used the synthetic visual features to train a classifier for unseen classes. What they differ in is the way to restrict synthetic visual features and the choice of visual features and semantic features. Zhu et al. \cite{Zhu18GAZSL} proposed to restrict synthetic visual features by adding a visual pivot regularization and they employed local visual features extracted from semantic regions of objects. By adding a classification penalty on synthetic visual features and using visual features from deeper CNN, Xian et al. \cite{Xian18FCLSWGAN} proposed a feature generation network. Li et al. \cite{li2019LiGAN} restricted synthetic visual features by adding multiple visual souls regularization. Different from them, we employ GAN to generate visual feature residual instead of visual feature.

\section{Methodology}
The definition of ZSL is as follows. Let $S = \left\{ \left( x_{n}, y_{n}, e \right) \mid x_{n} \in X^{s}, y_{n} \in Y^{s}, e \in E, n = 1, 2, \cdots, N \right\}$ be a training dataset, where $x_{n} \in \mathbf{R^{v}}$ is the visual feature of the $n$-th labeled image in the training dataset, $y_{n}$ is the class label of $x_{n}$, which belongs to seen-class set $Y^{s}$, $N$ is the number of samples, and $e \in \mathbf{R^{s}}$ is the semantic feature of a class in the total class set $Y$ which not only includes the seen-class set $Y^{s}$ but also includes the unseen-class set $Y^{u}$. Note that the unseen-class set $Y^{u}$ is disjoint with the seen-class set $Y^{s}$. Let $X$ represents the test image set, conventional ZSL is to learn a mapping $f: X \to Y^{u}$, while generalized ZSL is to learn a mapping $f: X \to Y$.

To tackle the ZSL problem, we propose a novel network to synthesize compact semantic visual features of unseen classes with adversarial feature residual, called AFRNet, and then train a classifier with these synthetic visual features and their corresponding labels for feature classification, the overall pipeline is shown in Figure 3. As shown in the feature generation phase, the AFRNet consists of three modules: residual generator, prototype predictor, and discriminator. The residual generator is used to generate the visual feature residual, and then the visual feature is synthesized by integrating the residual and the visual prototype predicted by the prototype predictor, the real visual features are extracted by an feature extractor (implemented by an existing categorization CNN). The discriminator tries to distinguish the synthetic visual features from the real visual features. Further, by applying a feature selection strategy to visual prototypes and real visual features, we could learn the compact semantic visual features. After the AFRNet is trained, as shown in the classification phase of Figure 3, the synthetic unseen-class visual features are used to train a classifier for feature classification.

In the following, firstly we introduce a semantically compact prototype predictor for predicting visual prototypes from semantic features. We then describe how to learn compact semantic visual feature from adversarial feature residual. Next, the employed classifier is introduced. Finally, the comparison of the proposed method to some related works is given.
\begin{figure*}[t]
\centering
\includegraphics[width=1.6 \columnwidth]{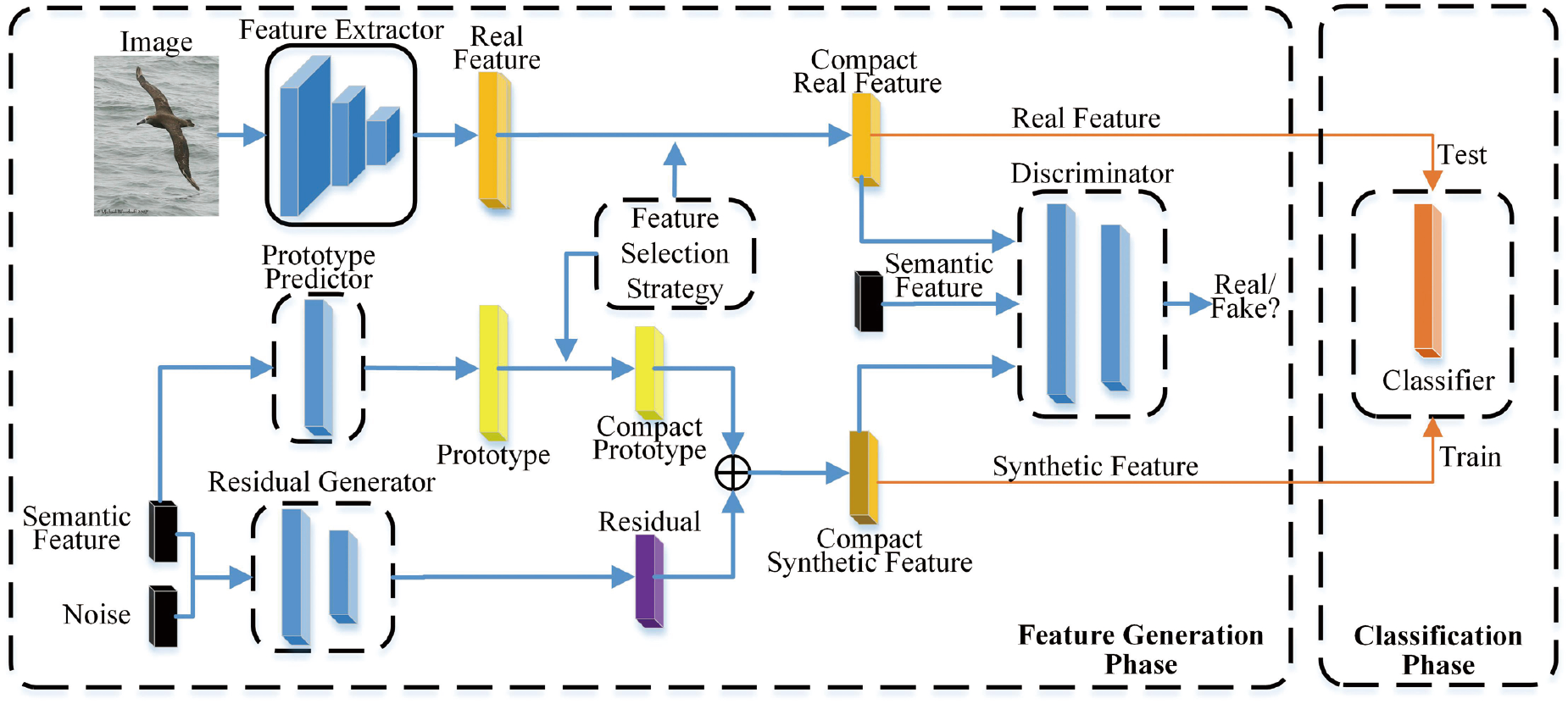}
\caption{Pipeline of the proposed method.}
\end{figure*}

\subsection{Semantically Compact Prototype Predictor}
\textbf{Predicting Visual Prototype.} Here, our goal is to learn a prediction function with a set of training data belonging to seen classes for predicting the visual prototypes of unseen classes from their semantic features. We use the mean vector of the visual feature vectors of each seen class as the visual prototype of each class. Suppose we have $N_{c}$ visual features for class $c$ in the training data, then the visual prototype $p_{c}$ for class $c$ is computed by $\frac{1}{N_{c}} \sum_{i=1}^{N_{c}} x_{c}^{i} $, where $x_{c}^{i} \in \mathbf{R^{v}}$ represents the $i$-th visual feature belonging to class $c$, $v$ is the dimensionality of a visual feature. We next denote semantic feature of class $c$ by $e_{c} \in \mathbf{R^{s}}$, where $s$ is the dimensionality of a semantic feature. After obtaining $C$ pairs of visual prototype and semantic feature $\left\{ \left( p_{c}, e_{c} \right) \mid c = 1, 2, \cdots, C \right\}$, for each dimension of visual prototype, we will train an individual SVR \cite{smola2004svr} with RBF kernel, the SVR used to predict the $j$-th dimension of the visual prototype is as follows:
\begin{equation}
\begin{split}
\min_{w^{j},\beta^{j},\bar{\beta^{j}},\delta} & \frac{1}{2}\left\| w^{j} \right\|^2 + \alpha\left( \frac{1}{C}\sum_{c=1}^{C}\left( \beta_{c}^{j} + \bar{\beta_{c}^{j}} \right)\right)\\
s.t. & (w^{j})^{T}\Phi^{j} \left( e_{c} \right) - p_{c}^{j} \leq \delta + \beta_{c}^{j} \\
& p_{c}^{j} - (w^{j})^{T}\Phi^{j}\left( e_{c} \right) \leq \delta + \bar{\beta_{c}^{j}} \\
& \beta_{c}^{j} \ge 0, \bar{\beta_{c}^{j}}\ge 0, \ c = 1, 2, \cdots, C
\end{split}
\end{equation}
where $\Phi^{j}\left( e_{c} \right)$ is the implicit semantic feature of class $c$  in kernel space, $w^{j}$ is trainable linear weight of the SVR. $p_{c}^{j}$ is the $j$-th dimension of the visual prototype of class $c$, $\delta$ is the margin, indicating any error less than it will not be counted, $\alpha$ is a penalty parameter. Note that each SVR takes $s$-dimension semantic feature as input and output $1$-dimension visual feature. Hence, $v$ SVRs can be trained independently so that they could be optimized parallelly to better capture relationship between semantic feature and every dimensions of visual prototype. Considering that the number of seen classes is probably smaller than the dimensionality of semantic feature, to avoid overfitting, the dimensionality of semantic feature will be firstly reduced before being fed into SVR. With the trained SVRs, given semantic features of seen classes and unseen classes, we predict visual prototypes of both of them. \\
\textbf{Predicting Compact Semantic Visual Prototype with Feature Selection Strategy.} The predicted visual prototypes can achieve a high performance only if visual features are consistent with semantic features. However, as mentioned before, visual features and semantic features are not inherently matching, so that it is inevitable that the predicted visual prototypes have error compared to the real prototypes. This error was ignored by previous methods, however, we argue that visual prototype dimension with smaller error is one that is better consistent with semantic feature. Hence, we propose a simple yet effective feature selection strategy which selects the $Top$-$K$ visual prototype dimensions with relatively smaller prediction errors to build a compact semantic visual prototype as:
\begin{equation}
\begin{split}
&\left[ j_{1},\cdots,j_{k},\cdots,j_{v} \right] = argsort \left[ \sum_{c=1}^{C} ( \Gamma^{j}( e_{c} ) - p^{j}_{c} )^2 \right] \\
&p_{c}^{\prime} = p_{c}\left[ j_{1},\cdots,j_{k} \right], \ c = 1,2,\cdots,C
\end{split}
\end{equation}
where $e_{c}$ is the semantic feature of class $c$, $\Gamma^{j}()$ is the SVR used to predict the $j$-th dimension of visual prototype, $p^{j}_{c}$ is the $j$-th dimension of visual prototype of class $c$, and $\left[ j_{1},\dots,j_{k},\cdots,j_{v} \right]$ is the index of visual prototype dimensions which rank in ascending order according to their prediction error, $p_{c}^{\prime}$ is the $Top$-$K$ dimensions of $p_{c}$ with smallest error, $K$ is a parameter which we fixed at $\frac{v}{2}$ in our experiments.

\subsection{Compact Visual Feature Learning from Adversarial Feature Residual}
Here, we describe how to learn compact semantic visual feature from adversarial feature residual in the proposed AFRNet. We employ the residual generator to generate visual feature residual, and then synthesize the compact semantic visual feature by combining the residual and the compact semantic visual prototype predicted by the aforementioned prototype predictor. The discriminator tries to distinguish the synthetic visual features from the real ones. After the adversarial training, we finally can synthesize compact semantic visual features. In this section, we begin with the general visual feature generation method as it is the basis of the proposed method, and then explain the proposed AFRNet in detail.

The general visual feature generation method employs the conditional WGAN \cite{arjovsky2017wgan} to generate visual features conditioned on their corresponding semantic features as:
\begin{equation}
\begin{split}
\min_{G} \max_{D} V = E \left[D\left(x,e\left(y\right)\right)\right]- E \left[D\left(\hat{x},e\left(y\right)\right)\right] \\
- \lambda E \left[\left(\left \| \nabla_{\bar{x}}D\left(\bar{x},e\left(y\right)\right) \right \|_{2}-1 \right)^2 \right]
\end{split}
\end{equation}
where $G$ and $D$ represent generator and discriminator respectively, which are both implemented by multi-layer perceptrons, $x$ is real visual feature, $\hat{x}=G\left(z,e\left(y\right)\right)$ is synthetic visual feature conditionally generated from noise $z$ and semantic feature $e\left(y\right)$ by generator $G$, $\bar{x} = \zeta x + \left(1-\zeta \right) \hat{x}$ with $\zeta \sim U\left(0,1\right)$ is used to estimate gradient. The objective of WGAN is to minimize Wasserstein distance which is implemented by the first two terms in Equation (3), and the last term is the gradient penalty of the discriminator, whose weight is controlled by a hyperparameter $\lambda$.

In the proposed AFRNet, the generator in conditional WGAN generates visual feature residuals instead of visual features. Specifically, given semantic feature $e_{y}$ and its corresponding compact semantic visual prototype $p_{y}^{\prime}$, we employ conditional WGAN to generate visual feature residual $r_{y}$ conditioned on $e_{y}$, then the compact semantic visual feature is synthesized by combining the residual $r_{y}$ with the compact semantic visual prototype $p_{y}^{\prime}$. The proposed method can be formalized as:
\begin{equation}
\begin{split}
\min_{G} \max_{D} V_{r} = & E \left[D\left(x,e(y)\right)\right] \\
& -E \left[D\left(r_{y}+p_{y}^{\prime},e(y)\right)\right] \\
& -\lambda E \left[\left(\left \| \nabla_{\bar{x}_{r}}D\left(\bar{x}_{r},e\left(y\right)\right) \right \|_{2}-1 \right)^2 \right]
\end{split}
\end{equation}
where $p_{y}^{\prime}$ is predicted by semantically compact prototype predictor, $r_{y}= G\left(z,e\left(y\right)\right)$ is generated by residual generator conditioned on its semantic feature. $(r_{y}+p_{y}^{\prime})$ is the synthetic compact semantic visual feature. $\bar{x}_{r} = \zeta x + \left(1-\zeta \right) \left(r_{y}+p_{y}^{\prime}\right)$ with $\zeta \sim U\left(0,1\right)$ is used to estimate gradient. The rest is similar to Equation (3).

After obtaining the synthetic visual features, the discriminator tries to distinguish the synthetic visual features from the real visual features. As the adversarial training goes, we end up with a powerful AFRNet which can synthesize compact semantic visual features that not only have less overlap but also are more consistent with semantic features.

\subsection{Classification}
Once the adversarial network has been trained, lots of visual features of unseen classes could be synthesized, associating with their labels. Then, ZSL is converted into a supervised classification problem. We could employ a naive softmax classifier as:
\begin{equation}
\begin{split}
\min_{\theta} L\left(\theta\right) = -\frac{1}{N} \sum_{i=1}^{N}log \ p\left( y_{i} \mid x_{i}; \theta \right)
\end{split}
\end{equation}
where $\theta$ is trainable linear transformation weight and $p( y_{i} \mid x_{i}; \theta )=\prod_{l=1}^{C} ( \frac{exp(\theta_{l}^{T}x_{i})}{\sum_{j=1}^{C} exp(\theta_{j}^{T}x_{i})} )^{1( y_{i}=l )}$. In testing phase, given a visual feature $x$, it is classified by:
\begin{equation}
f\left(x\right) = \mathop{\arg\max}_{y} p\left( y \mid x; \theta \right)
\end{equation}

\subsection{Comparison to Related Works}
Here, we compare the proposed method with the related works. Different from \cite{Zhu18GAZSL,Xian18FCLSWGAN,li2019LiGAN,paul2019SABR}, which all used GAN to generate visual features, we employ GAN to generate visual feature residuals, and then synthesize visual features by integrating the residuals and predicted visual prototypes. Since visual prototypes are explicitly predicted for both seen and unseen classes and the generated residuals are generally numerically much smaller than the distances among all the prototypes, the synthetic visual features are expected to be less overlapped for both seen and unseen classes.

Changpinyo et al. \cite{Changpinyo17EXEM} proposed to predict visual exemplars of unseen classes from semantic embeddings, and used the whole predicted visual exemplar either as visual training data or as ideal semantic embedding. Different from them, the proposed method adaptively selects some visual feature dimensions from the whole predicted visual prototype to build a compact semantic visual prototype, and applies this visual prototype to synthesize visual feature by being combined with visual feature residual.
\begin{table}[t]
	\centering
	\caption{Statistics of six ZSL datasets. Att = Attributes, TF = TF-IDF feature, SCS = SCS-split, SCE = SCE-split, PS = PS-split, S = Seen classes, U = Unseen classes, P = Part feature, R = Res101 feature.}
	\resizebox{.95\columnwidth}{!}{
		\begin{tabular}{cccccccccccc}
			\hline
			Dataset& Image& Att& TF& Feat& \multicolumn{2}{c}{SCS}&  \multicolumn{2}{c}{SCE}& \multicolumn{2}{c}{PS} \\
			\cline{6-7} \cline{8-9} \cline{10-11}
			& & & & & S& U& S& U& S& U \\
			\hline
			\hline
			CUB& 11,788& -& 7,551& P& 150& 50& 160& 40& -& -\\
			NAB& 49,562& -& 13,217& P& 323& 81& 323& 81& -& -\\
			APY& 15,339& 64& -& R& -& -& -& -& 20& 12\\
			AWA1& 37,475& 85& -& R& -& -& -& -& 40& 10\\
			AWA2& 37,322& 85& -& R& -& -& -& -& 40& 10\\
			SUN& 14,340& 102& -& R& -& -& -& -& 645& 72\\
			\hline
		\end{tabular}
	}
\end{table}

\section{Experimental Results}
\subsection{Experimental Setup}
\textbf{Datasets.} The proposed method is evaluated on the following six public datasets: Caltech USCD Birds-2011 (CUB) \cite{WahCUB_200_2011}, North America Birds (NAB) \cite{van2015NAB}, APascal-aYahoo (APY) \cite{farhadi2009APY}, Animals with Attributes (AWA1) \cite{Lampert14DAP}, renewed Animals with Attributes (AWA2) \cite{Xian17Comprehensive} and SUN attributes (SUN) \cite{patterson2012sun}. These datasets are of different scales and their statistics are summarized in Table 1. Note that CUB and NAB are two fine-grained datasets.\\
\textbf{Visual and Semantic Feature.} In order to make fair comparison, for APY, AWA1, AWA2 and SUN, as done in \cite{Xian17Comprehensive}, we use the 2048-D global features extracted by ResNet-101 \cite{He16resnet} which is pre-trained on ImageNet1000 as the visual features, and attributes as the semantic features. For the CUB and NAB, as done in \cite{elhoseiny2017ZSLPP,Zhu18GAZSL,Yu18SGA}, we use features which are merged with local features of several semantic regions of objects as the visual features, and Term Frequency-Inverse Document Frequency (TF-IDF) features as the semantic features. The TF-IDF features are commonly used for text description, which could represent the semantic feature of class-level text description. Specifically, the visual feature is 3584-D and 3072-D for CUB and NAB respectively, and we call these visual features extracted from local regions `part feature'. Following the previous works, we also transform the original TF-IDF feature to 200-D and 400-D via linear PCA operation for CUB and NAB respectively. The statistics of visual features and semantic features are reported in Table 1.\\
\textbf{Evaluation Protocol.} As most methods did, we evaluate the proposed method by computing average per-class $Top$-$1$ accuracy (ACC). In the conventional ZSL setting, we compute ACC of unseen classes. In the generalized ZSL setting, we compute ACCs of both seen classes and unseen classes and compute harmonic mean of seen and unseen accuracy. In addition, data split has a huge impact on performance. As suggested by \cite{elhoseiny2017ZSLPP}, on CUB and NAB, we evaluate the proposed method via SCS-split and SCE-split. Note that SCE-split is harder than SCS-split as the parent categories of unseen classes are exclusive to those of seen classes in SCE-split. Also note that few unseen classes in SCS-split have been seen by the pre-trained ImageNet1000 model, we use the same ImageNet1000 model as the other methods for fair comparison. Following \cite{Xian17Comprehensive}, we evaluate the proposed method on the APY, AWA1, AWA2 and SUN datasets with PS-split. The PS-split where unseen classes presented in ImageNet1000 are replaced with other classes is an improved version of the original SS-split. The detailed split information is reported in Table 1.\\
\textbf{Comparison Methods.} For comparison, we cite the results (reported in the corresponding papers) of fourteen existing methods on the APY, AWA1, AWA2 and SUN datasets, including DAP \cite{Lampert14DAP}, DEVISE \cite{Frome13DeViSE}, ALE \cite{Akata16ALE}, ESZSL \cite{romera2015ESZSL}, LATEM \cite{xian2016LATEM}, SSE \cite{zhang2015SSE}, SYNC \cite{Changpinyo16SYNC}, SAE \cite{Kodirov17SAE}, DEM \cite{zhang2017DEM}, GAZSL \cite{Zhu18GAZSL}, f-CLSWGAN \cite{Xian18FCLSWGAN}, SR-GAN \cite{Ye19SR-GAN}, SABR \cite{paul2019SABR}, LiGAN \cite{li2019LiGAN}. Similarly, we also list the results of seven existing methods on the fine-grained CUB and NAB datasets, including WAC-kernel \cite{elhoseiny2016WAC}, ESZSL \cite{romera2015ESZSL}, ZSLNS \cite{qiao2016ZSLNS}, SYNC \cite{Changpinyo16SYNC}, ZSLPP \cite{elhoseiny2017ZSLPP}, GAZSL \cite{Zhu18GAZSL}, SGA-DET \cite{Yu18SGA}.\\
\textbf{Implementation Details.} In the proposed method, prototype prediction is implemented by SVR with RBF kernel. The generator and discriminator are both three-layer MLP with ReLU activation, which both employ 4096 units in hidden layer. Hyper-parameters in WGAN are set as they are suggested by the author.

\subsection{Performance in Conventional ZSL Setting}
Since most state-of-the-art methods have been evaluated on APY, AWA1, AWA2 and SUN in the conventional ZSL setting, we first evaluate the proposed method on these datasets with PS-split and then compare it with fourteen state-of-the-art methods. All these methods are tested with Res101 features and attributes. Results are reported in Table 2. From Table 2, we can easily find out that the proposed method significantly outperforms all the existing methods. Specifically, the proposed method achieves an improvement about $12.2\%$ on APY, $4.4\%$ on AWA1, $8.0\%$ on AWA2, $1.2\%$ on SUN. The reason why the improvement on SUN is smaller than the others is probably that the visual prototypes on SUN are harder to be predicted due to the fact that it has more classes and less per-class images.
\begin{table}[t]
\centering
\caption{Comparative results (Top-1 accuracy) in the conventional ZSL setting on APY, AWA1, AWA2 and SUN.}
\resizebox{.8\columnwidth}{!}{
\begin{tabular}{ccccc}
\hline
Method& APY& AWA1& AWA2& SUN \\
\hline
DAP& 33.8& 44.1& 46.1& 39.9 \\
DEVISE& 39.8& 54.2& 59.7& 56.5 \\
ALE& 39.7& 59.9& 62.5& 58.1 \\
ESZSL& 38.3& 58.2& 58.6& 54.5 \\
LATEM& 35.2& 55.1& 55.8& 55.3 \\
SSE& 34.0& 60.1& 61.0& 51.5 \\
SYNC& 23.9& 54.0& 46.6& 56.3 \\
SAE& 8.3& 53.0& 54.1& 40.3 \\
DEM& 35.0& 68.4& 67.1& 61.9 \\
GAZSL& 41.1& 68.2& -& 61.3 \\
f-CSLWGAN& 40.5& 68.2& -& 60.8 \\
SR-GAN& 44.0& 72.0& -& 62.3 \\
SABR& -& -& 65.2& 62.8 \\
LiGAN& 43.1& 70.6& -& 61.7 \\
\hline
AFRNet(Ours)& \textbf{56.2}& \textbf{76.4}& \textbf{75.1}& \textbf{64.0} \\
\hline
\end{tabular}
}
\end{table}
\begin{table}[t]
	\centering
	\caption{Comparative results (Top-1 accuracy) in the conventional ZSL setting on the fine-grained CUB and NAB datasets.}
	\resizebox{.8\columnwidth}{!}{
		\begin{tabular}{ccccc}
			\hline
			Method& \multicolumn{2}{c}{CUB}&  \multicolumn{2}{c}{NAB} \\
			\cline{2-3} \cline{4-5}
			& SCS& SCE& SCS& SCE \\
			\hline
			\hline
			WAC-kernel& 33.5& 7.7& 11.4& 6.0 \\
			ESZSL & 28.5& 7.4& 24.3& 6.3 \\
			ZSLNS & 29.1& 7.3& 24.5& 6.8 \\
			SynC & 28.0& 8.6& 18.4& 3.8 \\
			ZSLPP & 37.2& 9.7& 30.3& 8.1 \\
			GAZSL & 43.7& 10.3& 35.6& 8.6 \\
			SGA-DET & 42.9& 10.9& 39.4& 9.7 \\
			\hline
			AFRNet(Ours) & \textbf{50.3}& \textbf{20.5}& \textbf{42.8}& \textbf{12.8} \\
			\hline
		\end{tabular}
	}
\end{table}
\begin{table*}[t]
	\centering
	\caption{Comparative results in the generalized ZSL setting on APY, AWA1, AWA2 and SUN. U = Top-1 accuracy of unseen classes, S = Top-1 accuracy of seen classes, H = Harmonic mean of unseen and seen classes accuracy.}
	\resizebox{1.6\columnwidth}{!}{
		\begin{tabular}{c|ccc|ccc|ccc|ccc}
			\hline
			Method& \multicolumn{3}{c}{APY}& \multicolumn{3}{c}{AWA1}& \multicolumn{3}{c}{AWA2}& \multicolumn{3}{c}{SUN} \\
			\cline{2-4} \cline{5-7} \cline{8-10} \cline{11-13}
			&  U&  S&  H&  U&  S&  H&  U&  S&  H&  U&  S&  H \\
			\hline
			\hline
			DAP& 4.8& 78.3& 9.0& 0.0& 88.7& 0.0& 0.0& 84.7& 0.0& 4.2& 25.1& 7.2 \\
			DEVISE& 4.9& 76.9& 9.2& 13.4& 68.7& 22.4& 17.1& 74.7& 27.8& 16.9& 27.4& 20.9 \\
			ALE& 4.6& 73.7& 8.7& 16.8& 76.1& 27.5& 14.0& 81.8& 23.9& 21.8& 33.1& 26.3 \\
			ESZSL& 2.4& 70.1& 4.6& 6.6& 75.6& 12.1& 5.9& 77.8& 11.0& 11.0& 27.9& 15.8 \\
			LATEM& 0.1& 73.0& 0.2& 7.3& 71.7& 13.3& 11.5& 77.3& 20.0& 14.7& 28.8& 19.5 \\
			SSE& 0.2& 78.9& 0.4& 7.0& 80.5& 12.9& 8.1& 82.5& 14.8& 2.1& 36.4& 4.0 \\
			SYNC& 7.4& 66.3& 13.3& 8.9& 87.3& 16.2& 10.1& 90.5& 18.0& 7.9& 43.3& 13.4 \\
			SAE& 0.4& 80.9& 0.9& 1.8& 77.1& 3.5& 1.1& 82.2& 2.2& 8.8& 18.0& 11.8 \\
			DEM& 11.1& 75.1& 19.4& 30.5& 86.4& 45.1& 30.5& 86.4& 45.1& 20.5& 34.3& 25.6 \\
			GAZSL& 14.2& 78.6& 24.0& 19.2& 86.5& 31.4& -& -& -& 21.7& 34.5& 26.7 \\
			f-CLSWGAN& 32.9& 61.7& 42.9& 57.9& 61.4& 59.6& -& -& -& 42.6& 36.6& 39.4 \\
			SR-GAN& 22.3& 78.4& 34.8& 41.5& 83.1& 55.3& -& -& -& 22.1& 38.3& 27.4 \\
			SABR& -& -& -& -& -& -& 30.3& 93.9& 46.9& 50.7& 35.1& \textbf{41.5} \\
			LiGAN& 34.3& 68.2& 45.7& 52.6& 76.3& 62.3& -& -& -& 42.9& 37.8& 40.2 \\
			\hline
			AFRNet(Ours)& 48.4& 75.1& \textbf{58.9}& 68.2& 69.4& \textbf{68.8}& 66.7& 73.8& \textbf{70.1}& 46.6& 37.6& \textbf{41.5} \\
			\hline
		\end{tabular}
	}
\end{table*}
\begin{table}[t]
	\centering
	\caption{Results with/without AFRNet-style feature generation method.}
	\resizebox{.7\columnwidth}{!}{
		\begin{tabular}{ccccc}
			\hline
			Method& \multicolumn{2}{c}{CUB}& \multicolumn{2}{c}{NAB} \\
			\cline{2-3} \cline{4-5}
			& SCS& SCE& SCS& SCE \\
			\hline
			AFRNet-non& 41.6& 9.1& 37.1& 5.7 \\
			AFRNet& 48.7& 18.6& 41.5& 12.7 \\
			\hline
		\end{tabular}
	}
\end{table}
\begin{table}[t]
	\centering
	\caption{Results with/without feature selection strategy. 1NN and AFR refer to 1NN classifier and AFRNet.}
	\resizebox{.95\columnwidth}{!}{
		\begin{tabular}{ccccccccc}
			\hline
			Method& \multicolumn{4}{c}{CUB}& \multicolumn{4}{c}{NAB} \\
			\cline{2-5} \cline{6-9}
			& \multicolumn{2}{c}{SCS}& \multicolumn{2}{c}{SCE}& \multicolumn{2}{c}{SCS}& \multicolumn{2}{c}{SCE} \\
			\cline{2-3} \cline{4-5} \cline{6-7} \cline{8-9}
			& 1NN& AFR& 1NN& AFR& 1NN& AFR& 1NN& AFR \\
			\hline
			w/o& 44.3& 48.7& 16.3& 18.6& 35.2& 41.5& 9.4& 12.7 \\
			w& 48.7& 50.3& 18.2& 20.5& 38.2& 42.8& 9.8& 12.8 \\
			\hline
		\end{tabular}
	}
\end{table}

For a more detailed evaluation, we also test the proposed method on two fine-grained datasets, CUB and NAB. We then compare it with seven recent state-of-the-art methods. All these methods are tested using part features and TF-IDF features. To make evaluation more challenging, we evaluate these methods with both easier SCS-split and harder SCE-split. Table 3 shows us the results. As shown in Table 3, the proposed method outperforms all the competitors with a significant performance gain. Specifically, on CUB, we achieve performance gain $6.6\%$ and $9.6\%$ under SCS-split and SCE-split. Significantly, the accuracy under SCE-split ($20.5\%$) is about $90\%$ higher than that of previous state-of-the-art ($10.9\%$) on CUB. We visualize features of 10 unseen classes from CUB as shown in Figure 2 B, both the accuracy gain and the visualization indicate that the AFRNet can generate visual features with less overlap. The gain on NAB is slightly smaller than on CUB, which is $3.4\%$ and $3.1\%$ under SCS-split and SCE-split, this is probably because NAB is a larger dataset with 404 categories, which means inter-class difference on NAB is relatively subtle.\\

\subsection{Performance in Generalized ZSL Setting}
We evaluate the proposed method on APY, AWA1, AWA2 and SUN with PS-split in the generalized ZSL setting. Then, we conduct comparison with fourteen state-of-the-art methods. All these methods are evaluated using Res101 features and attributes. Results are reported in Table 4. Similar to results in the conventional ZSL setting, the proposed method achieves a significantly better performance than previous methods: $58.9\%$ vs $45.7\%$ on APY, $68.8\%$ vs $62.3\%$ on AWA1, $70.1\%$ vs $46.9\%$ on AWA2. On SUB, the proposed method only reaches the state-of-the-art performance probably because visual prototypes are harder to be predicted on SUB. In addition, accuracy of seen classes and that of unseen classes are better balanced in the proposed method than other methods, which informs us the proposed method has a better generalization to unseen classes. From Table 2 to Table 4, we also note that all the GAN-based methods consistently outperform other methods. Among all the GAN-based methods, the proposed method achieves the best performance, this indicates that the AFRNet method and feature selection strategy are effective to improve ZSL, which we will detailedly analysis in the Ablation Study section.

\subsection{Ablation Study}
\textbf{Effect of Feature Generation Method.} To prove benefit of the feature generation method proposed in the AFRNet, comparison experiments are conducted on CUB and NAB under both SCS-split and SCE-split using AFRNet-style feature generation method (AFRNet method) and non-AFRNet-style method (AFRNet-non method). The results are reported in Table 5. It is obvious that AFRNet method achieves a significant performance gain against AFRNet-non method: $7.1\%$ and $9.5\%$ under SCS-split and SCE-split on CUB; $4.4\%$ and $7.0\%$ under SCS-split and SCE-split on NAB. Notably, the accuracy of AFRNet method is more than $100\%$ higher than that of AFRNet-non method under SCE-split on both CUB and NAB. This gain indicates that the AFRNet method can generate visual features with less overlap as shown in Figure 2 B, and these features could be used to train a more generalizable classifier. Note that both AFRNet method and AFRNet-non method in Table 5 have not employed feature selection strategy. \\
\textbf{Effect of Feature Selection Strategy.} To demonstrate benefit of the feature selection strategy (FSS), we conduct evaluation on both CUB and NAB under both SCS-split and SCE-split using both naive 1NN classifier and AFRNet. Results with or without FSS are reported in Table 6. We note that methods with FSS achieve better performance than methods without it whatever the evaluation settings are. This tells us that FSS is able to select visual feature dimensions that are better consistent with semantic features.\\

\section{Conclusion}
We propose a novel adversarial network called AFRNet to learn compact semantic visual features for ZSL. Unlike existing feature generation methods, the proposed AFRNet generates visual feature residual, and then synthesizes the visual feature by integrating the residual with the predicted visual prototype. Consequently, the synthetic visual features are less overlapped and classifier trained on these features is more generalizable. In addition, on the basis of existing prototype prediction method, we propose a novel feature selection strategy which can adaptively select semantically consistent visual feature elements from the original visual feature. The proposed method is proved to outperform existing state-of-the-art methods with a significant improvement by extensive experimental results on six benchmarks datasets.

\section{ Acknowledgments}
This work was supported by the Strategic Priority Research Program of the Chinese Academy of Sciences (XDB32070100) and National Natural Science Foundation of China (U1805264, 61421004, 61573359). We thank the anonymous reviewers so much for their helpful comments and suggestions.
\bibliography{AAAI-LiuL.3879}
\bibliographystyle{aaai}
\end{document}